\documentclass[11pt,a4paper]{article}
\usepackage[hyperref]{acl2018}
\usepackage{times}
\usepackage{latexsym}

\usepackage{url}

\usepackage{natbib}
\usepackage{verbatim}
\usepackage{amsmath}
\usepackage{cases}
\usepackage{cleveref}
\usepackage{soul}
\usepackage{pgfplots}
\usetikzlibrary{pgfplots.groupplots}
\pgfplotsset{compat=1.14}
\usepackage[colorinlistoftodos,prependcaption,textsize=tiny]{todonotes}
\usepackage{todonotes}

\newcommand{\eupdt}[1]{#1}
\newcommand{\note}[2][1=]{}

\newcommand{\cancel}[1]{}

\aclfinalcopy %
\def\aclpaperid{797} %

\title{Restricted Recurrent Neural Tensor Networks: Exploiting 
Word Frequency and Compositionality} 

\author{Alexandre Salle$^{1}$ \quad Aline Villavicencio$^{1,2}$ \\
  $^{1}$Institute of Informatics, Federal University of Rio Grande do Sul (Brazil) \\
  $^{2}$School of Computer Science and Electronic Engineering, University of Essex (UK)\\
  {\tt alex@alexsalle.com \quad avillavicencio@inf.ufrgs.br }}

\date{}

\begin{document}

\maketitle

\begin{abstract}

Increasing the capacity of recurrent neural networks (RNN) usually involves  augmenting the size of the hidden layer,  with significant increase of computational cost. Recurrent neural tensor networks (RNTN) increase capacity using  distinct hidden layer weights for each word, but with greater costs in memory usage. In this paper, we introduce restricted recurrent neural tensor networks (r-RNTN) which reserve distinct hidden layer weights for frequent vocabulary words while sharing a single set of weights for infrequent words. Perplexity evaluations show that for fixed hidden layer sizes, r-RNTNs improve language model performance over RNNs using only a small fraction of the parameters of unrestricted RNTNs. These results hold for r-RNTNs using Gated Recurrent Units and Long Short-Term Memory.\footnote{This is a preprint of the paper that will be presented at the 56th Annual Meeting of the Association for Computational Linguistics.}

\end{abstract}

\section{Introduction}

\eupdt{Recurrent}  neural networks (RNN), which compute their next output conditioned on a previously stored hidden state, are a natural solution to sequence modeling.
\citet{Mikolov2010} applied RNNs to word-level language modeling (we refer to \eupdt{this model as s-RNN}), outperforming traditional n-gram methods. 
\eupdt{However, increasing} capacity (number of tunable parameters)
\eupdt{by augmenting} the size $H$ of the hidden (or recurrent) layer ---
\eupdt{to model more complex distributions ---} 
results in a significant increase in computational cost, which is $O(H^2)$.

\citet{Sutskever2011} increased the performance of %
\eupdt{a} character-level language model 
\eupdt{with} a multiplicative RNN (m-RNN), the factored approximation of a recurrent neural tensor network (RNTN), which maps each symbol to separate hidden layer weights (referred to as recurrence matrices from hereon). Besides increasing model capacity while keeping computation constant, this approach has another motivation: viewing the RNN's hidden state as being transformed by each new symbol in the sequence, it is intuitive that different symbols will transform \eupdt{the network's hidden} state in different ways \cite{Sutskever2011}. \eupdt{Various studies on \eupdt{compositionality 
similarly }argue that some words are better modeled by matrices than by vectors \citep{Baroni2010,Socher2012}.} %
Unfortunately, having separate recurrence matrices for each symbol requires memory that is linear in the symbol vocabulary size ($|V|$). This is not an issue for character-level models, which have small vocabularies, but is prohibitive for word-level models which can have vocabulary size in the millions if we consider surface forms.  

In this paper, we propose the Restricted RNTN (r-RNTN) which uses only
$K < |V|$ recurrence matrices. Given that $|V|$ words must be assigned $K$ matrices, we map the most frequent $K-1$ words to the first $K-1$ matrices, and share the $K$-th matrix among the remaining words.
\eupdt{This mapping is driven by the statistical intuition that frequent words are more likely to appear in diverse contexts and so require richer modeling, }
\eupdt{and by the greater presence of predicates and function words among the most frequent words in standard \eupdt{corpora like COCA \cite{davies2009385+}.}} 
\eupdt{As a result, adding $K$ matrices to the s-RNN both} increases model capacity and 
\eupdt{satisfies the idea that } 
\eupdt{some words are better represented by matrices.} 
Results show that r-RNTNs improve language model performance over s-RNNs even for small $K$ with no computational overhead, and even for small $K$ approximate the performance of RNTNs using a fraction of the parameters.
\eupdt{We also experiment with r-RNTNs using Gated Recurrent Units (GRU) \citep{Cho2014LearningPR} and Long Short-Term Memory (LSTM) \citep{Hochreiter1997LongSM}, obtaining lower perplexity for fixed hidden layer sizes.}
\eupdt{This paper discusses related work ($\S$\ref{sec:related}), and presents r-RNTNs ($\S$\ref{sec:r-RNTN}) along with the evaluation method ($\S$\ref{sec:materials}). We conclude with results ($\S$\ref{sec:results}), and suggestions for future work.}

\section{Related Work}
\label{sec:related}
We focus \eupdt{
on} related work that addresses language modeling via RNNs, word representation, and conditional computation.

Given a sequence of words $(x_1, ..., x_T)$, a language model gives the probability $P(x_t | x_{1 \dots t-1})$ for $t \in [1, T]$. Using a RNN, \citet{Mikolov2010} created the s-RNN language model %
given by:
\begin{align}
\label{eq:srnnh}
h_t = \sigma (W_h x_t + U_h h_{t-1} + b_h) \\
P(x_t | x_{1 \dots t-1}) = x_t^T Softmax(W_o h_t + b_o) 
\end{align}
where $h_t$ is the hidden state \eupdt{represented by a vector of dimension $H$}, $\sigma(z)$ is the \eupdt{pointwise} logistic function, $W_h$ is the $H \times V$ embedding matrix, $U_h$ is the $H \times H$ recurrence matrix, \eupdt{ $W_o$ is the $V \times H$ output matrix}, and $b_h$ and $b_o$ are bias terms. \eupdt{
Computation} is $O(H^2)$, so increasing model capacity by increasing $H$ quickly becomes intractable.

The RNTN proposed by \citet{Sutskever2011} is nearly identical to the s-RNN, but the recurrence matrix in \cref{eq:srnnh} is replaced by a tensor as follows:
\begin{equation}
h_t = \sigma (W_h x_t + U_h^{i(x_t)} h_{t-1} + b_h)
\label{eq:rntnh}
\end{equation}
where $i(z)$ maps a hot-one encoded vector to its integer representation. \eupdt{Thus the $U_h$ tensor is} %
composed of $|V|$ recurrence matrices, and at each step of sequence processing the matrix corresponding to the current input is used to transform the hidden state. 
\eupdt{
The authors also} proposed m-RNN, a factorization of the $U_h^{i(x_t)}$ \eupdt{matrix} \eupdt{into $U_{lh} diag(v_{x_t}) U_{rh}$} to reduce the number of parameters\eupdt{, where $v_{x_t}$ is a \emph{factor} vector of the current input $x_t$}, but like the RNTN, memory still grows linearly with $|V|$.
The RNTN has the property that input symbols have both a vector representation given by the embedding and a matrix representation given by the recurrence matrix, unlike the s-RNN where symbols are limited to vector representations. 

The integration of both vector and matrix representations has been discussed 
\eupdt{
but with a focus on representation learning and not sequence modeling \cite{Baroni2010,Socher2012}.} \eupdt{For instance, \citet{Baroni2010} argue for nouns to be represented as vectors and \eupdt{adjectives as matrices.}} %

\citet{Irsoy2014} used m-RNNs for the task of sentiment classification 
\eupdt{and obtained equal or better performance than s-RNNs}. 
Methods that use conditional computation \cite{cho2014exponentially, bengio2015conditional,shazeer2017outrageously} are similar to RNTNs and r-RNTNs, but rather than use a static mapping, these methods train gating functions which do the mapping. 
Although these methods can potentially learn better policies than our method,
they are significantly more complex, requiring the use of reinforcement learning \citep{cho2014exponentially,bengio2015conditional} or additional loss functions \citep{shazeer2017outrageously}, and more linguistically opaque (one must learn to interpret the mapping performed by the gating functions). 

\eupdt{Whereas our work is concerned with updating the network's hidden state, \citet{chen2015strategies} introduce a technique that better approximates the output layer's Softmax function by allocating more parameters to frequent words.}

\section{Restricted Recurrent Neural Tensor Networks}
\label{sec:r-RNTN}
\eupdt{To balance expressiveness and computational cost, we} 
propose restricting the size of the recurrence tensor in the RNTN such that memory does not grow linearly with vocabulary size, while still keeping dedicated matrix representations for a subset of words in the vocabulary. We call these Restricted Recurrent Neural Tensor Networks (r-RNTN), which modify 
\cref{eq:rntnh} as follows:
\begin{equation}
h_t = \sigma (W_h x_t + U_h^{f(i(x_t))} h_{t-1} + b_h^{f(i(x_t))})
\label{eq:rrntnh}
\end{equation}
where $U_h$ is a tensor of $K < |V|$ matrices of size $H \times H$, $b_h$ is a $H \times K$ bias matrix with 
\eupdt{columns} 
indexed by $f$. The function $f(w)$ maps each vocabulary word to an integer between $1$ and $K$. 

We use the following definition for $f$: 
\begin{equation}
f(w) = min(rank(w), K)
\end{equation}
where $rank(w)$ is the rank of word $w$ when the vocabulary is sorted by decreasing order of unigram frequency. 

This is an intuitive choice because words which appear more often in the corpus tend to have more variable contexts, so it makes sense to dedicate a large part of model capacity to them. A second argument is that frequent words tend to be predicates and function words.
We can imagine that predicates and function words transform the meaning of the current hidden state of the RNN through matrix multiplication, whereas nouns, for example, add meaning through vector addition, \eupdt{following \citet{Baroni2010}}.

\eupdt{We also perform initial experiments with r-RNTNs using LSTM and GRUs. A GRU is described by}
\begin{align}
r_t &= \sigma (W_h^r x_t + U_h^r h_{t-1} + b_h^r) \\
z_t &= \sigma (W_h^z x_t + U_h^z h_{t-1} + b_h^z) \\
\noindent \tilde{h}_t &= \tanh (W_h^h x_t + U_h^h (r_t \odot h_{t-1}) + b_h^h) \\
h_t &= z_t \odot h_{t-1} + (1 - z_t) \odot \tilde{h}_t
\end{align}
and an LSTM by
\begin{align}
f_t &= \sigma (W_h^f x_t + U_h^f h_{t-1} + b_h^f) \\
i_t &= \sigma (W_h^ix_t + U_h^i h_{t-1} + b_h^i) \\
o_t &= \sigma (W_h^o x_t + U_h^o h_{t-1} + b_h^o) \\
\tilde{c}_t &= \tanh (W_h^c x_t + U_h^c h_{t-1} + b_h^c) \\
c_t &= i_t \odot \tilde{c}_t + f_t \odot c_{t-1} \\
h_t &= o_t \odot tanh(c_t)
\end{align}
We create r-RNTN GRUs (r-GRU) by making $U_h^h$ and $b_h^h$ input-specific (as done in \cref{eq:rrntnh}). For r-RNTN LSTMs (r-LSTM), we do the same for $U_h^c$ and $b_h^c$.

\pgfplotsset{every axis/.append style={
                    axis x line=middle,    %
                    axis y line=middle,    %
                    xlabel={$x$},          %
                    ylabel={$y$},          %
                    label style={font=\tiny},
                    tick label style={font=\tiny}  
                    }}

\begin{figure}
\begin{tikzpicture}
\pgfplotsset{
    every non boxed x axis/.style={},
    every non boxed y axis/.append style={y axis line style=-},
    legend style={nodes={scale=0.6, transform shape}}
}
 \begin{groupplot}[
    group style={
        group name=my fancy plots,
        group size=1 by 2,
        xticklabels at=edge bottom,
        vertical sep=0pt
    },
    width=8.0cm,
    xmin=0, xmax=10000,
    enlargelimits=false
]

\nextgroupplot[
               ytick={130,140,150,160},
               axis x line=none, 
               axis y discontinuity=crunch,
                   xmode=log,
	ylabel=Test PPL,
    ymin=120,ymax=160,width=8.0cm,height=4.35cm
     ]
               ]
\addplot[color=blue,mark=x] coordinates {
	(1, 146.671939)
    (2, 146.108378)
    (5, 143.642528)
    (10, 139.515340)
    (20, 135.626202)
    (30, 135.920162)
    (40, 134.088379)
    (50, 137.381475)
    (60, 133.868577)
    (70, 133.715856)
    (80, 131.660600)
    (90, 133.390431)
    (100, 131.170104)
    (1000, 131.262480)
    (3000, 130.401893)
    (5000, 127.066613)
    (7000, 127.082184)
    (10000, 128.791481)
};
\addlegendentry{r-RNTN $f$}

\addplot[color=red,mark=+] coordinates {
	(1, 146.671939)
    (2, 148.522653)
    (5, 142.207827)
    (10, 141.353293)
    (20, 143.800175)
    (30, 141.091965)
    (40, 139.926902)
    (50, 139.597592)
    (60, 139.868643)
    (70, 140.149847)
    (80, 139.348723)
    (90, 138.311409)
    (100, 138.927813)
    (1000, 134.035191)
    (3000, 130.357768)
    (5000, 128.282031)
    (7000, 128.378844)
    (10000, 128.791481)
};
\addlegendentry{r-RNTN $f_{mod}$}

\addplot[color=purple,mark=none] coordinates {
	(1,  128.791481)
    (2,  128.791481)
    (5,  128.791481)
    (10,  128.791481)
    (20,  128.791481)
    (30,  128.791481)
    (40,  128.791481)
    (50,  128.791481)
    (60,  128.791481)
    (70,  128.791481)
    (80,  128.791481)
    (90,  128.791481)
    (100,  128.791481)
    (1000,  128.791481)
    (3000,  128.791481)
    (5000,  128.791481)
    (7000,  128.791481)
    (10000,  128.791481)
};
\addlegendentry{RNTN}

\addplot[dotted,color=black,mark=none] coordinates {
	(1, 146.671939)
    (2, 146.671939)
    (5, 146.671939)
    (10, 146.671939)
    (20, 146.671939)
    (30, 146.671939)
    (40, 146.671939)
    (50, 146.671939)
    (60, 146.671939)
    (70, 146.671939)
    (80, 146.671939)
    (90, 146.671939)
    (100, 146.671939)
    (1000, 146.671939)
    (3000, 146.671939)
    (5000, 146.671939)
    (7000, 146.671939)
    (10000, 146.671939)
};
\addlegendentry{s-RNN $H = 100$}

\addplot[dashed,color=green,mark=none] coordinates {
	(1, 133.676746)
    (2, 133.676746)
    (5, 133.676746)
    (10, 133.676746)
    (20, 133.676746)
    (30, 133.676746)
    (40, 133.676746)
    (50, 133.676746)
    (60, 133.676746)
    (70, 133.676746)
    (80, 133.676746)
    (90, 133.676746)
	(100, 133.676746)
    (1000, 133.676746)
    (3000, 133.676746)
    (5000, 133.676746)
    (7000, 133.676746)
    (10000, 133.676746)
};
\addlegendentry{s-RNN $H = 150$}

\nextgroupplot[
               ytick={87,90,93},
               axis y line=middle,
                   xmode=log,
	xlabel=K,
	ylabel=,
    xmin=0,xmax=10000,
    ymin=85,ymax=93,width=8.0cm,height=2.75cm
     ]
\addplot[color=black,mark=+] coordinates {
(1, 92.155138)    
(10, 89.648645)   
(20, 88.17794)    
(30, 87.566263)   
(40, 87.936876)   
(50, 87.735845)   
(60, 87.37506)    
(70, 88.203061)   
(80, 87.589533)   
(90, 87.583393)   
(100, 87.528083)  
};
\addlegendentry{r-GRU}

\addplot[color=purple,mark=x] coordinates {
(1, 88.671459)  
(10, 87.530516) 
(20, 87.34944)  
(30, 88.214871) 
(40, 87.973033) 
(50, 87.551759) 
(60, 87.114482) 
(70, 87.423518) 
(80, 86.566493) 
(90, 86.890301) 
(100, 87.070865)
};
\addlegendentry{r-LSTM}
\end{groupplot}
\end{tikzpicture}
\caption{\eupdt{PTB test PPL as $K$ varies from $1$ to $10 000$ ($100$ for gated networks). At $K=100$, the r-RNTN with $f$ mapping already closely approximates the much bigger RNTN, with little gain for bigger $K$, showing that dedicated matrices should be reserved for frequent words as hypothesized.}}
\label{fig:keffectlarge}
\end{figure}
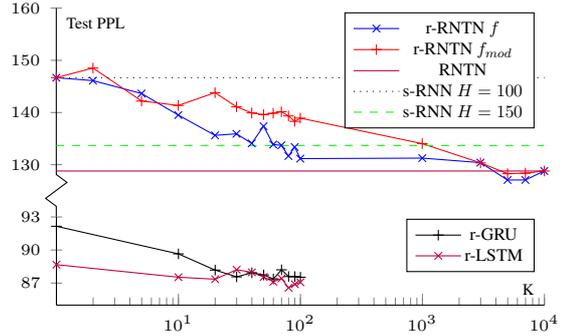

\begin{table*}[t]
  \centering
  \setlength\tabcolsep{4pt} %
  \begin{tabular}{|c|c|cc|cc||c|c|cc|}
\hline
& & \multicolumn{2}{|c|}{PTB} & \multicolumn{2}{|c||}{text8} & & & \multicolumn{2}{|c|}{PTB} \\
Method & $H$ & \# Params & Test PPL & \# Params & Test PPL & Method & $H$ & \# Params & Test PPL \\
\hline
s-RNN & 100 & 2M & 146.7 & 7.6M & 236.4 & GRU & 244 & 9.6M & 92.2  \\
r-RNTN $f$ & 100 & 3M & 131.2 & 11.4M & 190.1 & GRU & 650 & 15.5M & 90.3  \\ 
RNTN & 100 & 103M & 128.8 & 388M & - & r-GRU $f$ & 244 & 15.5M & \textbf{87.5} \\ 
\cline{7-10}
m-RNN & 100 & 3M & 164.2 & 11.4M & 895.0  & LSTM & 254 & 10M & 88.8 \\
s-RNN & 150 & 3M & 133.7 & 11.4M & 207.9 & LSTM & 650 & 16.4M & \textbf{84.6} \\
r-RNTN $f$ & 150 & 5.3M & \textbf{126.4} & 19.8M & \textbf{171.7} & r-LSTM $f$ & 254 & 16.4M & 87.1 \\
\hline
  \end{tabular}
\caption{\eupdt{Comparison of validation and test set perplexity for r-RNTNs with $f$ mapping ($K=100$ for PTB, $K = 376$ for text8) versus s-RNNs and m-RNN. r-RNTNs with the same $H$ as corresponding s-RNNs significantly increase model capacity and performance with no computational cost. The RNTN was not run on text8 due to the number of parameters required.}}
  \label{tab:keffect}
\end{table*}
\section{Materials}
\label{sec:materials}
We evaluate 
\eupdt{s-RNNs, RNTNs, and r-RNTNs by training and}
measuring model perplexity (PPL)
on the Penn Treebank (PTB) corpus \cite{Marcus1994} using the same preprocessing as \citet{Mikolov2011}. \eupdt{Vocabulary size is 10000.}
\eupdt{}

For an r-RNTN with $H = 100$, we vary the tensor size $K$ from $1$, which corresponds to the s-RNN, all the way up to $10000$, which corresponds to the unrestricted RNTN. As a simple way to evaluate our choice of rank-based mapping function $f$, we compare it to %
a pseudo-random variant:
\begin{equation}
f_{mod}(w) = rank(w) \bmod K
\label{eq:fmod}
\end{equation}

We also compare results to 1) \eupdt{an} s-RNN with $H = 150$, which has the same number of parameters as an r-RNTN with $H = 100$ and $K = 100$. \eupdt{2) An m-RNN with $H = 100$ with the size of factor vectors set to $100$ to match this same number of parameters. \eupdt{3)} An additional r-RNTN with $H=150$ is trained to show that performance scales with $H$ as well.}

We split each sentence into 20 word sub-sequences and run \eupdt{stochastic gradient descent} via backpropagation through time 
for 20 steps \eupdt{without mini-batching}, only reseting the RNN's hidden state between sentences. 
\eupdt{Initial learning rate (LR) is $0.1$ and halved when the ratio of the validation perplexity between successive epochs is less than $1.003$,
stopping training when validation improvement drops below this ratio for $5$ \eupdt{consecutive} epochs.} 
\eupdt{We use Dropout \cite{Srivastava2014} with $p = .5$ on the softmax input to reduce overfitting.}
\eupdt{Weights are drawn from $\mathcal{N}(0, .001)$; gradients are not clipped.}
\eupdt{To validate our proposed method, we also evaluate r-RNTNs using the much larger text8\footnote{http://mattmahoney.net/dc/textdata.html} corpus
with a 90MB-5MB-5MB train-validation-test split, mapping words which appear less than 10 times to \emph{$\langle unk \rangle$} for a total vocabulary size of $37751$.}

\eupdt{Finally, we train GRUs, LSTMs, and their r-RNTN variants using the PTB corpus and parameters similar to those used by \citet{Zaremba2014}. All networks use embeddings of size $650$ and a single hidden layer. Targeting $K = 100$, we set $H = 244$ for the r-GRU and compare \eupdt{with} a GRU with $H = 650$ which has the same \eupdt{number} of parameters. The r-LSTM has $H=254$ to match the \eupdt{number} of parameters \eupdt{of} an LSTM with $H=650$. The training procedure is the same 
\eupdt{as above}
but \eupdt{with} mini-batches of size 20, 35 timestep sequences without state resets, initial LR of 1, Dropout on all non-recurrent connections, weights drawn from $\mathcal{U}(-.05,.05)$, and gradients norm-clipped to $5$.}
 
\section{Results}
\label{sec:results}
\eupdt{Results \eupdt{are shown} in \cref{fig:keffectlarge} and \cref{tab:keffect}.}

Comparing the r-RNTN to the baseline s-RNN with $H = 100$ \eupdt{(\cref{fig:keffectlarge})}, %
as model capacity grows with K, test set perplexity drops, showing that r-RNTN is an effective way to increase model capacity with no additional computational cost. As expected, the $f$ mapping outperforms \eupdt{the 
baseline} $f_{mod}$ mapping at smaller $K$. As $K$ increases, we see a convergence of both mappings. 
This \eupdt{may be due to}
matrix sharing at large $K$ between frequent and %
infrequent words because of the modulus operation in \cref{eq:fmod}.
As infrequent words are rarely observed, 
frequent words dominate the matrix updates and approximate having distinct matrices, as they would have with the $f$ mapping. 

It is remarkable that even with $K$ \eupdt{as} small as $100$, the r-RNTN approaches the performance of the RNTN with a small fraction of the parameters. This reinforces our hypothesis that complex transformation modeling afforded by distinct matrices is needed for frequent words, but not so much for infrequent words which can be well represented by a shared matrix and a distinct vector embedding. As shown in \cref{tab:keffect}, with an equal number of parameters, the r-RNTN with $f$ mapping outperforms the s-RNN with a bigger hidden layer. It appears that heuristically allocating increased model capacity as done by the $f$ based r-RNTN is a better way to increase performance than simply increasing hidden layer size, which also incurs a computational penalty.

\eupdt{Although m-RNNs have been successfully employed in character-level language models with small vocabularies, they are seldom used in word-level models. The poor results shown in \cref{tab:keffect} could explain why.\footnote{It should be noted that \citet{Sutskever2011} suggest m-RNNs would be better optimized using second-order gradient descent methods, whereas we employed only first-order gradients in all models we trained to make a fair comparison.}}

\eupdt{For fixed hidden layer sizes, r-RNTNs yield significant improvements to s-RNNs, GRUs, and LSTMs, confirming the advantages of distinct representations.}

\section{Conclusion and Future Work}
\label{sec:conclusions}
In this paper, \eupdt{we proposed restricted recurrent neural tensor networks, a model that restricts the size of recurrent neural tensor networks by mapping frequent words to distinct matrices and infrequent words to shared matrices. r-RNTNs were} motivated by the need to increase RNN model capacity without increasing computational costs, while also satisfying the %
ideas \eupdt{that some words are better modeled by matrices rather than vectors \cite{Baroni2010,Socher2012}}. We achieved both goals by pruning the size of the recurrent neural tensor network described by \citet{Sutskever2011} via sensible word-to-matrix mapping. Results validated our hypothesis that frequent words \eupdt{benefit from} richer, dedicated modeling 
\eupdt{as reflected in large perplexity improvements for low values of $K$}.

Interestingly, results for s-RNNs and r-GRUs suggest that given the \emph{same number of parameters}, it is possible to obtain higher performance by increasing $K$ and reducing $H$. This is not the case with r-LSTMs, perhaps to due to our choice of which of the recurrence matrices to make input-specific. We will further investigate both of these phenomena in future work, experimenting with different combinations of word-specific matrices for r-GRUs and r-LSTMs (rather than only $U_h^h$ and $U_h^c$), and combining our method with recent improvements to gated networks in language modeling \cite{jozefowicz2016exploring,merity2018regularizing,melis2018on} which we believe are orthogonal and hopefully complementary to our own. 

Finally, we plan to compare frequency-based and additional, linguistically motivated $f$ mappings (for example different inflections of a verb sharing a single matrix) with mappings learned via conditional computing to measure how external linguistic knowledge contrasts with knowledge automatically inferred from training data.

\section*{Acknowledgments}
This research was partly supported by CAPES and CNPq (projects 312114/2015-0, 423843/2016-8, and 140402/2018-7).

\bibliographystyle{acl_natbib.bst}
\bibliography{bib_manual}

\appendix

\end{document}